\documentclass[letterpaper, 10 pt, journal, twoside]{IEEEtran}
\usepackage{graphics} 
\usepackage{epsfig} 
\usepackage{mathptmx} 
\usepackage{times} 
\usepackage{amsmath} 
\usepackage{amssymb}  
\usepackage{textcomp}
\usepackage{algorithm}
\usepackage{algpseudocode}
\usepackage{cite}
\usepackage{subfigure}
\usepackage{cancel}
\usepackage{ulem}
\usepackage{xcolor}
\usepackage{booktabs}
\usepackage{multirow}

\ifCLASSINFOpdf
\else
\fi
\usepackage{tikz}
\usepackage{textcomp}
\usepackage{hyperref}
\usepackage{lipsum}

\newcommand\copyrighttext{\centering {\fontsize{6.6}{10}\selectfont 2377-3766 (c) 2019 IEEE. Personal use is permitted, but republication/redistribution requires IEEE permission. See http://www.ieee.org/publications\_standards/publications/rights/index.html for more information.}
}

\newcommand\copyrighttexttop{\centering {\fontsize{6.6}{10}\selectfont This article has been accepted for publication in a future issue of this journal, but has not been fully edited. Content may change prior to final publication. Citation information: DOI 10.1109/LRA.2020.2967324, IEEE Robotics and Automation Letters}}

\newcommand\copyrightnotice{%
\begin{tikzpicture}[remember picture,overlay]
\node[anchor=south,yshift=10pt] at (current page.south) 
{\parbox{\dimexpr\textwidth+32\fboxsep\relax}{\copyrighttext}};
\end{tikzpicture}%
}

\newcommand\copyrightnoticetop{%
\begin{tikzpicture}[remember picture,overlay]
\node[anchor=north,yshift=-2pt] at (current page.north) {\parbox{\dimexpr\textwidth+28\fboxsep\relax}{\copyrighttexttop}};
\end{tikzpicture}%
}

\begin{document}
%
\title{A Hybrid Compact Neural Architecture for\\Visual Place Recognition}

%
%
%

\author{
Marvin Chanc\'an$^{1,3}$, Luis Hernandez-Nunez$^{2,3}$, Ajay Narendra$^{4}$, Andrew B. Barron$^{4}$, and Michael Milford$^{1}$
\thanks{Manuscript received: September 5, 2019; Revised December 1, 2019; Accepted December 27, 2019.}
\thanks{This paper was recommended for publication by Editor Xinyu Liu upon evaluation of the Associate Editor and Reviewers' comments.
This work was supported by the Peruvian Ministry of Education to M. Chanc\'an and by an ARC Future Fellow FT140101229 to M. Milford.} 
\thanks{$^{1}$School of Electrical Engineering and Computer Science, Queensland University of Technology, Brisbane, QLD 4000, Australia
        }%
\thanks{$^{2}$Center for Brain Science \& Department of Physics, Harvard University, Cambridge, MA 02138, USA
        }%
\thanks{$^{3}$School of Mechatronics Engineering, Universidad Nacional de Ingenier\'ia, Lima, R\'imac 15333, Peru {\tt\small mchancanl@uni.pe}}%
\thanks{$^{4}$Department of Biological Sciences, Macquarie University, Sydney, NSW 2109, Australia}
\thanks{Digital Object Identifier (DOI): see top of this page.}
}
%
%

\markboth{IEEE Robotics and Automation Letters. Preprint Version. Accepted December, 2019}
{Chanc\'an \MakeLowercase{\textit{et al.}}: A Hybrid Compact Neural Architecture for Visual Place Recognition} 

%



\maketitle

\begin{abstract}

State-of-the-art algorithms for visual place recognition, and related visual navigation systems, can be broadly split into two categories: computer-science-oriented models including deep learning or image retrieval-based techniques with minimal biological plausibility, and neuroscience-oriented dynamical networks that model temporal properties underlying spatial navigation in the brain. In this letter, we propose a new compact and high-performing place recognition model that bridges this divide for the first time. Our approach comprises two key neural models of these categories: (1) \textit{FlyNet}, a compact, sparse two-layer neural network inspired by brain architectures of fruit flies, \textit{Drosophila melanogaster}, and (2) a one-dimensional continuous attractor neural network (CANN). The resulting \textit{FlyNet+CANN} network incorporates the compact pattern recognition capabilities of our FlyNet model with the powerful temporal filtering capabilities of an equally compact CANN, replicating entirely in a hybrid neural implementation the functionality that yields high performance in algorithmic localization approaches like SeqSLAM. We evaluate our model, and compare it to three state-of-the-art methods, on two benchmark real-world datasets with small viewpoint variations and extreme environmental changes -- achieving 87\% AUC results under day to night transitions compared to 60\% for Multi-Process Fusion, 46\% for LoST-X and 1\% for SeqSLAM, while being 6.5, 310, and 1.5 times faster, respectively. 
\end{abstract}


\begin{IEEEkeywords}
Biomimetics, Localization, Visual-Based Navigation
\end{IEEEkeywords}

%

\section{Introduction}\label{sec:introduction}
%
%
%
%
\IEEEPARstart{P}{erforming} visual place recognition (VPR) reliably is a challenge for any robotic system or autonomous vehicle operating over long periods of time in real-world environments. This is mainly due to a range of visual appearance changes over time (e.g. day/night or weather/seasonal cycles), viewpoint variations or even perceptual aliasing (e.g. multiple places may look similar) \cite{Lowry2016}. Convolutional neural networks (CNN), heavily used in a range of computer vision tasks \cite{Lecun2015}, have also been applied to the field of VPR with great success over the past five years \cite{zetao2014, zetao2017}; typically only used in real-time with dedicated hardware (GPU) though \cite{n2015,n2015-2, xin2018}. However, as vanilla CNN models, trained on benchmark datasets such as ImageNet \cite{imagenet_cvpr09} or Places365 \cite{places365}, generally neglect any temporal information between consecutive images. Conversely, sequence-based algorithms such as SeqSLAM \cite{seqslam} are often applied on top of these models to achieve state-of-the-art results on VPR tasks by matching two or more sequences of images.  \copyrightnoticetop \copyrightnotice

Related research in visual navigation has recently used computer-science-oriented recurrent neural networks (RNN) \cite{lstm} in an attempt to model the multi-scale spatial representation and network dynamics found in the entorhinal cortex of mammalian brains \cite{Banino2018,cueva2018}. While the results are promising, these systems are tested only in small synthetic environments, and the integration of neuroscience-oriented recurrent models such as continuous attractor neural networks (CANN) \cite{pi_cogmap, comp_model_gc} is not well explored. Only recently, analytic theories to unify both types of recurrent networks, trained on navigation tasks, have been proposed \cite{unify}.

\begin{figure}[!t]
   \centering
   \includegraphics[width=\columnwidth, height=45mm]{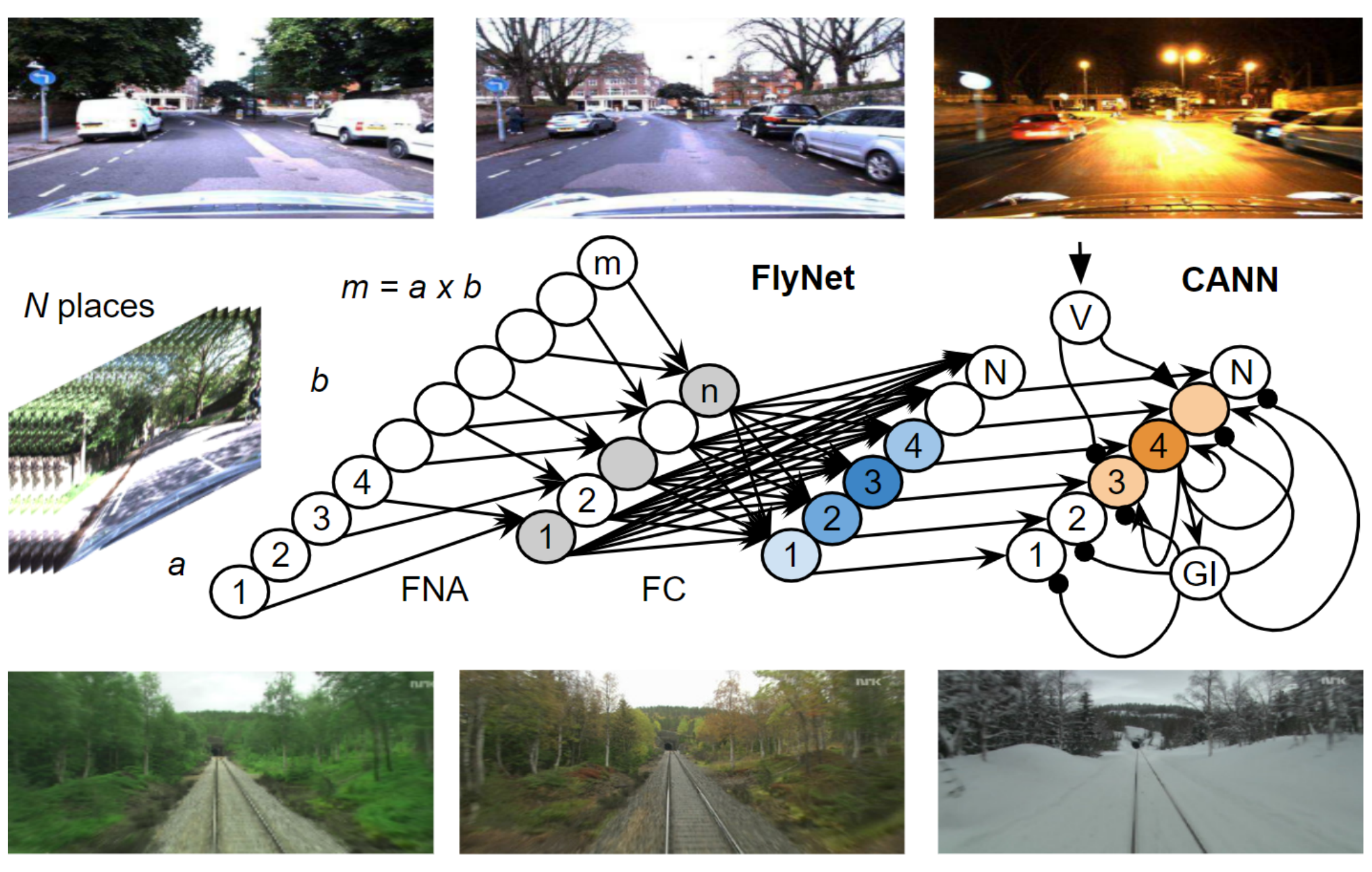}
   \vspace{-6mm}
   \caption{\textbf{FlyNet+CANN hybrid neural architecture}. Our FlyNet model comprises a hidden layer inspired by the \textit{Drosophila} olfactory neural circuit, FlyNet algorithm (FNA), and a fully connected (FC) output layer. We integrate FlyNet with a continuous attractor neural network (CANN) to perform appearance-invariant visual place recognition. Experiments on two real-world datasets, Oxford RobotCar (top) and Nordland (bottom), show that our hybrid model achieves competitive results compared to conventional approaches, but with a fraction of computational footprint (see Fig. \ref{model_size_1}).}
   \label{fly_algo}
   \vspace{-2mm}
\end{figure}

\begin{figure}[!t]
   \centering
   \includegraphics[width=\columnwidth]{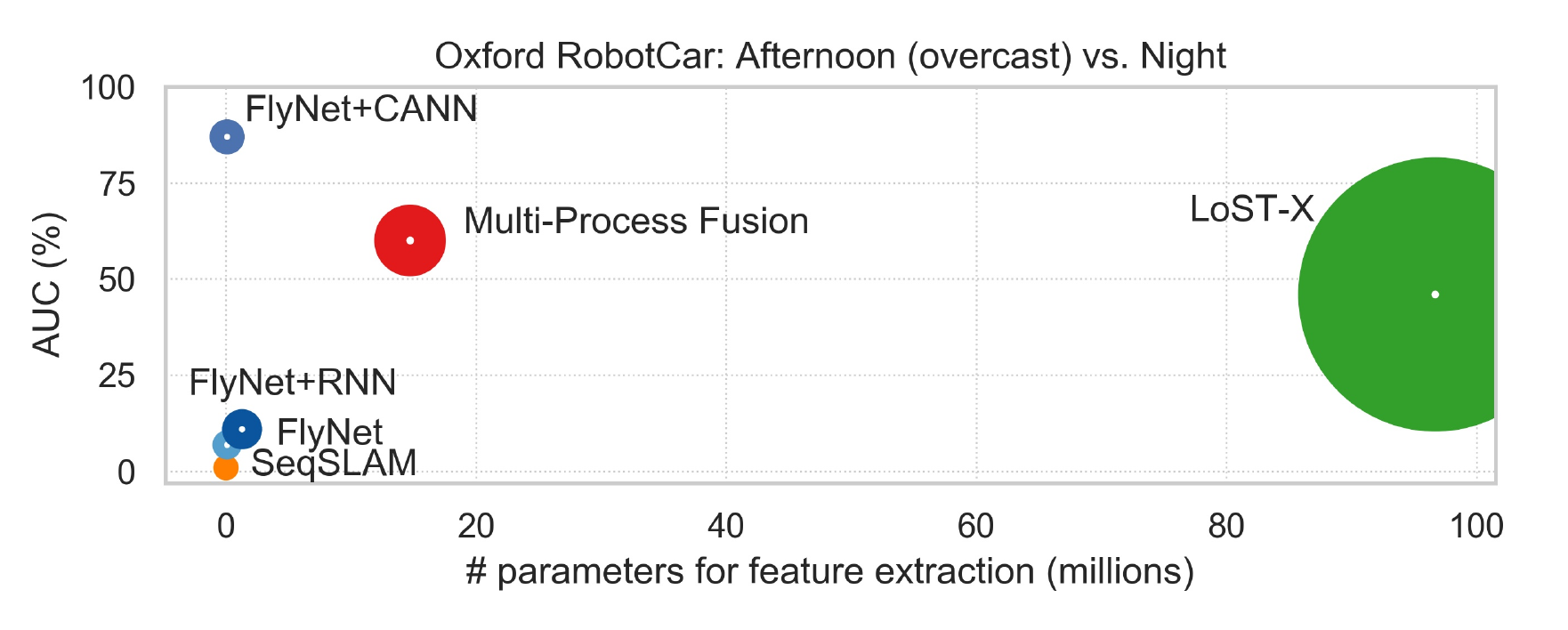}
   \vspace{-8mm}
   \caption{\textbf{Oxford RobotCar AUC performance vs. Network Size}. Footprint comparison for the most challenging appearance change (day to night).}
   \label{model_size_1}
   \vspace{-4mm}
\end{figure}

In this work, we propose a hybrid neural network that incorporates both computer-science- and neuroscience-oriented models, as in recent work \cite{tianjic,dashnet}, but for VPR tasks for the first time\footnote{Project page: {\tt mchancan.github.io/projects/FlyNet}}. Our approach comprises two key components (see Fig. \ref{fly_algo}): FlyNet, a compact neural network inspired by the \textit{Drosophila} olfactory neural circuit, and a 1-\textit{d} CANN as our temporal model that encodes sequences of images to perform appearance-invariant VPR using real data. The resulting FlyNet+CANN model achieves competitive AUC results on two benchmark datasets, but with far less parameters, minimal training time and smaller computational footprint than conventional deep learning and algorithmic-based approaches. In Fig. \ref{model_size_1}, for instance, the area of the circle is proportional to the number of layers per model, being 213 for the ResNet-based LoST-X pipeline \cite{garg2018lost}, 13 for Multi-Process Fusion \cite{mpf}, and 3 for our proposed FlyNet+CANN hybrid model.

The rest of the paper is structured as follows. Section \ref{sec:related} provides a brief overview of VPR research and the biological inspiration for our hybrid neural architecture; Section \ref{sec:methods} describes the FlyNet model in detail; Sections \ref{sec:experiments} and \ref{sec:results} present the experiments and results, respectively, where we compare our approach to three state-of-the-art VPR methods; and Section \ref{sec:conclusion} provides discussion around our biologically-inspired model as well as future work.

%


\section{Related Work}\label{sec:related}

This section outlines some key biological background for navigation in insect and mammalian brains, reviews the use of deep-learning-based approaches for VPR, and discusses recent developments in temporal filtering techniques for sequential data to further improve performance.

\subsection{Navigation in Biological Brains and Robots}

Our understanding of how animals navigate using vision has been used as an inspiration for designing effective localization, mapping and navigation algorithms. RatSLAM \cite{ratslam2004} is one example of this, using a model based on the rodent brain to perform visual SLAM over large real-world environments for the first time \cite{ratslam2008}. Likewise, researchers have developed a range of robotic navigation models based on other animals including insects \cite{copegreen, webbcricket, antbot}.


Insects such as ants, bees and flies exhibit great capabilities to navigate \cite{c24,c30, c3, c31,c34}. In fact, their brains share the same general structure \cite{c3}, \cite{c36}, with the central complex being closely associated with navigation, orientation and spatial learning \cite{c39,c40}. Place recognition is, however, most likely mediated by processing within the \textit{mushroom bodies} (MB), a separate pair of structures within their brains that are known to be involved in classification, learning, and recognition of both olfactory and visual information in bees and ants \cite{c39}. They receive densely coded and highly processed input from the sensory lobes, which then connects \textit{sparsely} to a large number of intrinsic neurons within the MB. Their structure has been likened to a \textit{multi-layer perceptron} (MLP) and considered optimal for learning and classifying complex input \cite{c49}.

These impressive capabilities, achieved with relatively small brains, make them attractive models for roboticists. For FlyNet, we take inspiration from algorithmic insights found in the fruit fly olfactory neural circuit. Our focus here is primarily on taking high-level inspiration from the size and structure of the fly brain and investigating the extent to which it can be integrated with recurrent-based networks for VPR tasks, much as in the early RatSLAM work and related development \cite{neuroslam}.

\subsection{Deep Neural Networks for Visual Place Recognition}

CNN models have been applied to a range of image recognition tasks, including VPR, with great success across many challenging real-world datasets with both visual appearance and viewpoint changes \cite{netvlad, noh2017, tori2015, garg2018lost}, and large-scale problems \cite{6dof, deepdsair}. Despite their success, these approaches often rely on the use of CNN models pre-trained on various computer vision datasets \cite{Long_2015_CVPR,n2015,n2015-2, netvlad}. Training these models in an end-to-end fashion specifically for VPR has also recently been proposed \cite{netvlad, zetao2017, zetao2018}. However, they are still using common network architectures, e.g. AlexNet \cite{alex2012}, VGG \cite{vgg} or ResNet \cite{resnet2016}, with slight changes to perform VPR tasks. All these systems share common undesirable characteristics with respect to their widespread deployability on real robots including large network sizes, extensive computing, and training requirements. In contrast, we propose the usage of compact neural models such as FlyNet to alleviate these requirements, while leveraging the temporal information found in most VPR datasets by using an equally compact CANN model.

\subsection{Modeling Temporal Relationships}\label{sec:temporal}

To access and exploit the power of temporal information in many applications, researchers have developed a range of RNN including long short-term memory (LSTM) \cite{lstm}. These temporal-based approaches have been applied specifically to visual navigation \cite{Banino2018} and spatial localization \cite{cueva2018} in artificial agents. In a nice closure back to the inspiring biology, these approaches led to the emergence of grid-like representations, among other cell types found in mammalian brains \cite{moser2008}, when training RNN cells to perform path integration \cite{pi_cogmap} and navigation \cite{unify}. RatSLAM \cite{ratslam2004}, one of the older approaches to filtering temporal information in a neural network, incorporated multi-dimensional CANN models with pre-assigned weights and structure set up to model the neural activity dynamics of place and grid cells found in the rat mammalian brain. Other non-neural techniques have been developed including SeqSLAM \cite{seqslam} in order to match sequences of pre-processed frames to provide an estimate of place, with a range of subsequent works \cite{naseerflows, newmanexperience, smart, lipatchtracker}. \copyrightnoticetop \copyrightnotice

The work to date has captured many key aspects of the VPR problem, investigating complex but powerful deep learning-based approaches, bio-inspired models that work in simulation or small laboratory mazes, and mammalian-brain based models with competitive real-world robotics performance. In this letter, we attempt to merge the desirable properties of several of these computer-science- and neuroscience-oriented models by developing a new bio-inspired, hybrid neural network for VPR tasks based on insect brain architectures such as FlyNet, which is extremely compact and can incorporate the filtering capabilities of a 1-\textit{d} CANN to achieve competitive localization results. We also show how our compact FlyNet model can easily be adapted to other temporal filtering techniques including SeqSLAM and RNN.

\section{Method Overview}\label{sec:methods}

We briefly describe recent development inspired by fruit fly brains such as the \textit{fly algorithm} \cite{dasgupta2017}. We then present our FlyNet algorithm (FNA) inspired by the \textit{fly algorithm}, and propose our single-frame, multi-frame, and hybrid models.

\subsection{Fly Algorithm} \label{flyalgo}

Recent research in brain-inspired computing suggests that \textit{Drosophila} olfactory neural circuits identify odors by assigning similar neural activity patterns to similar input odors \cite{dasgupta2017, cluster_insect}. These small brain cells perform a three-step procedure as the input odor goes through a three-layer neural circuit \cite{dasgupta2017}. First, the firing rates across the first layer are centered to the same mean for all odors (removing the odor concentration dependence). Second, a binary, sparse random matrix connects the second layer to the third layer, where each neuron receives and sums about 10\% of the firing rates from the second layer. Third, through a winner-take-all (WTA) circuit, only the highest-firing 5\% neurons across the third layer are used to generate a specific binary tag of the input odor.

The \textit{fly algorithm} is then proposed in \cite{dasgupta2017} to mimic the pattern recognition capabilities found in the fly brain, at a broad level and from a functional computer science perspective. Being mathematically defined as a binary locality-sensitive hash (LSH) function; a new class of LSH algorithms (see Eq. \ref{lsh}) but with relevant differences such as requiring significantly fewer computations as it uses sparse, binary random projections instead of dense, Gaussian random projections typical in LSH functions \cite{lsh}.

\begin{equation}
    Pr[h(p)=h(q)]=sim(p,q)
    \label{lsh}
\end{equation}
where $sim(p,q)$ is the similarity function, and $h:\mathbb{R}^m \to \mathbb{Z}^n$ is the LSH function if for any $p,\, q\in\mathbb{R}^m$, $Pr$ is $sim(p,q) \in [0,1]$.


\subsection{Proposed FlyNet Algorithm}\label{fna}

\begin{figure}
   \centering
   \includegraphics[width=0.7\columnwidth]{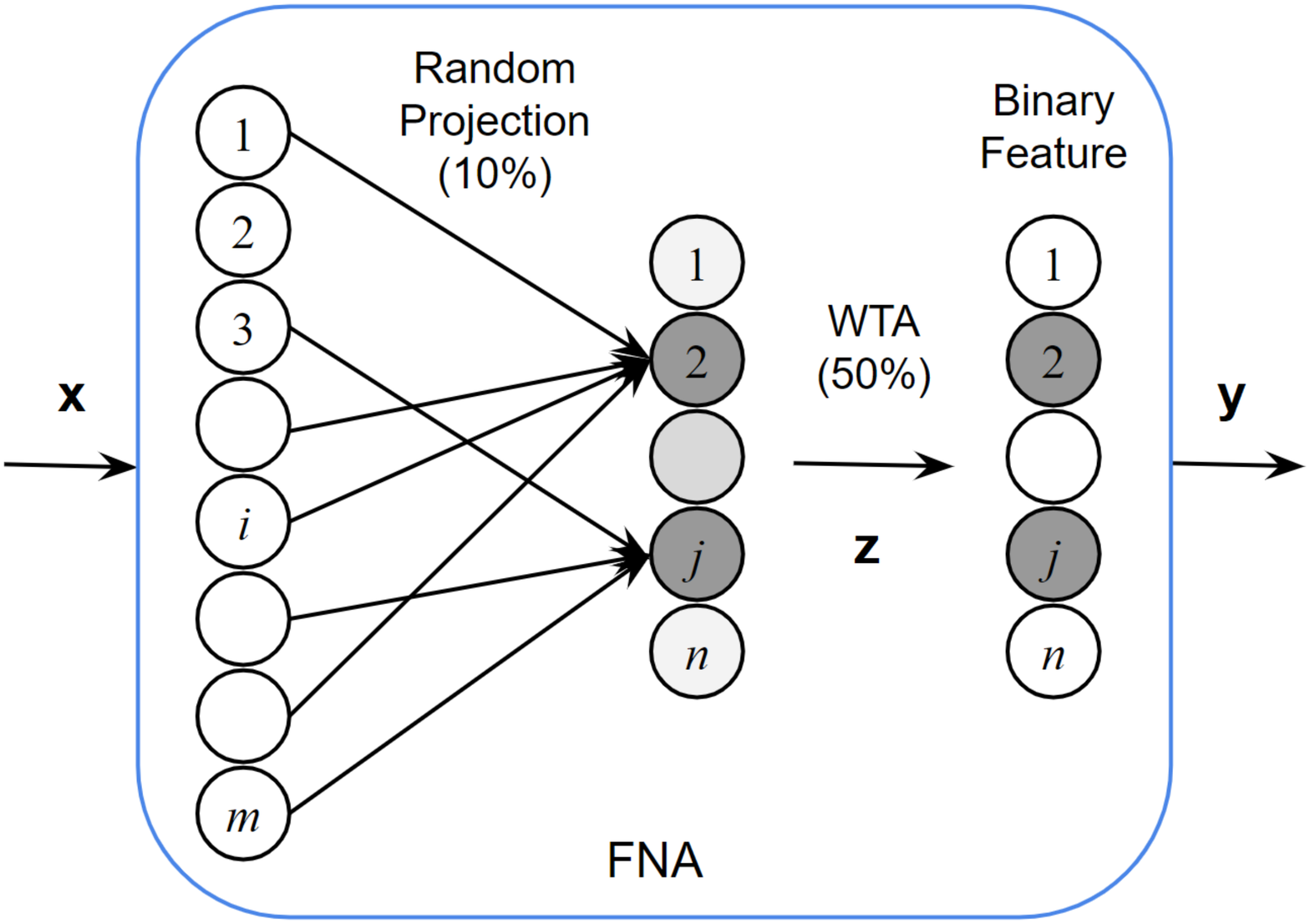}
   \caption{\textbf{FNA mapping}. The \textit{random projection} here shows only the connections to $z_2$ and $z_j$ within the second layer, but all the units in that layer connect with 10\% of the input units $x_i$.}
   \label{fna_pic}
\end{figure}

We leverage the \textit{fly algorithm} from a computer vision perspective to propose our FlyNet algorithm (FNA), see Algorithm 1. The FNA mapping, shown in Fig. \ref{fna_pic}, uses a sampling ratio of 10\% within the first layer, similar to the \textit{fly algorithm}. A WTA circuit of 50\% (instead of 5\% as in the \textit{fly algorithm}), is then used to generate a binary, compact output representation of our input image. Additional details on the choice and sensitivity of these parameters are provided in Section \ref{sec:fn_vs_single}. We also perform an image preprocessing step, to obtain $\mathbf{x}$, before applying Algorithm 1. Details on this procedure are outlined in Section \ref{sec:datasets}.



\begin{algorithm}
\caption[a]{FlyNet Algorithm (FNA)}
\hspace*{\algorithmicindent} \textbf{Input:} $\mathbf{x}\in \mathbb{R}^{m}$ \\
\hspace*{\algorithmicindent} \textbf{Output: $\mathbf{y}$ $\in \mathbb{Z}^{n}$}, $n<m$
\begin{algorithmic}[1]
\State Initialize $\mathbf{W}\in\mathbb{Z}^{n \times m}$: A binary, sparse random connection matrix between the input $\mathbf{x}$ and the output $\mathbf{y}$.
\State Compute the output $\mathbf{y} = \mathbf{W}\mathbf{x}$: Each output $y_j$ receives and sums 10\% randomly selected input values $x_i$.
\State WTA circuit: Set the top 50\% output values $y_i$ to 1, and the remaining to 0.
\end{algorithmic}
\end{algorithm}

\subsection{FlyNet-based Models}\label{sec:baselines}

\begin{figure}[!h]
   \centering
   \includegraphics[width=0.85\columnwidth]{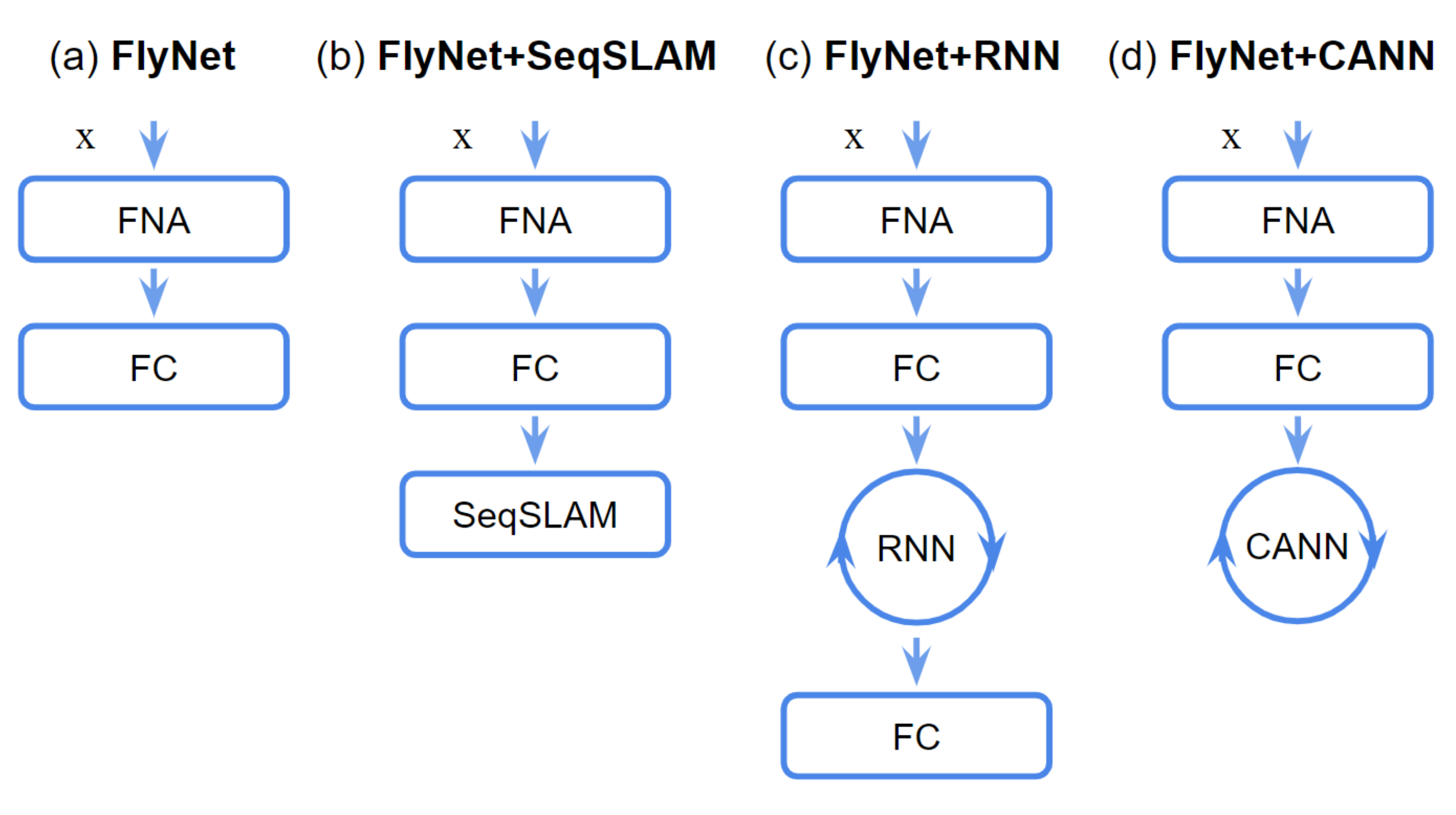}
   \vspace{-2mm}
   \caption{\textbf{FlyNet baselines}. Proposed (a) single-frame and (b, c, d) multi-frame models including the (d) hybrid FlyNet+CANN neural network.}
   \label{flynet}
\end{figure}

We implement a range of VPR models that leverage the FNA compact representations, including one single-frame model, and three multi-frame models with temporal filtering capabilities, see Fig. \ref{flynet}.

\subsubsection{\textbf{FlyNet}}

The FlyNet model, shown in Fig. \ref{flynet} (a), is our bio-inspired two-layer neural network that comprises the FNA as a hidden layer, and a fully connected (FC) output layer. We configure FlyNet to have a gray-scale input image dimension of $m=32\times 64$, and an output dimension of $n=64$. The FNA output $\mathbf{y}$ then feeds into a 1000-way \textit{linear} MLP which computes a particular class score for each input image. 

\subsubsection{\textbf{FlyNet+SeqSLAM}}

We incorporate the SeqSLAM algorithm \cite{seqslam} on top of our single-frame FlyNet network, as per previous research described in Sections \ref{sec:introduction} and \ref{sec:related}, see Fig. \ref{flynet} (b). The resulting model is a multi-frame baseline which we can compare along with our other temporal filtering-based models FlyNet+RNN and FlyNet+CANN.

\subsubsection{\textbf{FlyNet+RNN}}
It is a purely neural model that incorporates a vanilla RNN on top of FlyNet for temporal information processing, see Fig. \ref{flynet} (c). We also investigated the use of other types of RNN such as gated recurrent units (GRU) and LSTM. However, they showed no significant performance improvements despite having far more parameters. \copyrightnoticetop \copyrightnotice

\subsubsection{\textbf{FlyNet+CANN}}

It is our hybrid and also a purely neural model for sequence-based VPR tasks, see Fig. \ref{flynet} (d). We implemented a variation of the CANN architecture introduced in the RatSLAM work \cite{ratslam2008}, but using a 1-\textit{d} CANN model proposed in \cite{cann_model}, motivated by its suitability as a compact neural network-based way to implement the filtering capabilities of SeqSLAM \cite{seqslam}. As described in Section \ref{sec:temporal}, a CANN is a type of recurrent network that utilizes pre-assigned weights within its configuration. In Fig. \ref{fly_algo} (middle) we show our detailed FlyNet+CANN implementation where, in contrast to an RNN, a unit within the CANN layer can excite or inhibit itself and units nearby using excitatory (arrows) or inhibitory (rounds) connections, respectively, and can also include a global inhibitor (GI) unit in its main structure. For this implementation, activity shifts in our 1-\textit{d} CANN model, representing movement through the environment, were implemented with a direct shift and copy action. Although this could be implemented with more biologically faithful details such as velocity (V) units and asymmetric connections, as in prior CANN research \cite{strattonspike}.


\section{Experiments}\label{sec:experiments}


To evaluate the capabilities of our proposed FlyNet-based models, we conduct extensive experiments on two of the most widespread benchmarks used in VPR, the Nordland \cite{n2013} and Oxford RobotCar \cite{robotcar} datasets. We compare FlyNet (alone) with other related single-frame VPR methods and neural networks. Furthermore, we also compare our hybrid, multi-frame neural network to three state-of-the-art, multi-frame VPR approaches: SeqSLAM \cite{seqslam}, LoST-X \cite{garg2018lost}, and Multi-Process Fusion (MPF) \cite{mpf}. In this section, we describe in detail these network configurations and dataset preparation.

\subsection{Real-World Datasets}\label{sec:datasets}

\subsubsection{\textbf{Nordland}}

The Nordland dataset, introduced in \cite{n2013} for VPR research, comprises four single traverses of a train journey, in northern Norway, including extreme seasonal changes across spring, summer, fall, and winter. This dataset is primarily used to evaluate generalization over visual appearance changes, as instantiated through its four-season coverage. In our experiments, we use three traverses to perform VPR at 1 fps as in \cite{n2013}. We particularly use the summer traversal for training, and the remaining for testing, see Table \ref{table_datasets}.

\subsubsection{\textbf{Oxford RobotCar}}

The Oxford RobotCar dataset \cite{robotcar} provides over 100 traverses with different lighting (e.g. day, night) and weather (e.g. direct sun, overcast) conditions through a car ride in Oxford city; which implicitly contains various challenges of pose and occlusions such as pedestrians, vehicles, and bicycles for instance. In our evaluations, we use the same subsets as in \cite{garg2018lost} with overcast (autumn) for training, and day/night for testing, see Table \ref{table_datasets}.

\begin{table}[!ht]
\caption{Sequence-Based Datasets for VPR (reference/query)}
\label{table_datasets}
\vspace{-2mm}
\begin{center}
\begin{tabular}{llll}
\toprule
\textbf{Dataset} & \textbf{Appearance Changes} & \textbf{Viewpoint Changes} \\
\midrule
\multirow{2}{*}{Nordland Railway} & Small (summer/fall) &  \multirow{2}{*}{Small} \\
 & Extreme (summer/winter) &  \\
\midrule
\multirow{2}{*}{Oxford RobotCar} & Small (overcast/day) & \multirow{2}{*}{Moderate} \\
 & Extreme (overcast/night) &  \\
\bottomrule
\end{tabular}
\end{center}
\vspace{-4mm}
\end{table}


\textbf{Data Preprocessing.} In all our experiments, we use a sequence of 1000 images per traversal (reference or query) and provide full resolution RGB images to all the models, being $1920\times1080$ for Nordland and $1280\times960$ for Oxford RobotCar. Our FlyNet baselines convert the images into single-channel (gray-scale) frames normalized between [0, 1], and then resize them to $32\times64$. While the state-of-the-art methods apply their default image preprocessing before feeding their models.

\subsection{Evaluation Metrics}

We evaluate the VPR performance of our models using precision-recall (PR) curves and area under the curve (AUC) metrics. The tolerance used to consider a query place as a correct match is being within 20 frames around the ground truth location for the Nordland dataset, and up to 50 meters (10 frames) away from the ground truth for the Oxford RobotCar dataset, as per previous research \cite{garg2018lost}, \cite{mpf}, \cite{mao2019}.

\begin{figure*}[!t]
   \centering
   \subfigure{
   \includegraphics[width=0.98\columnwidth]{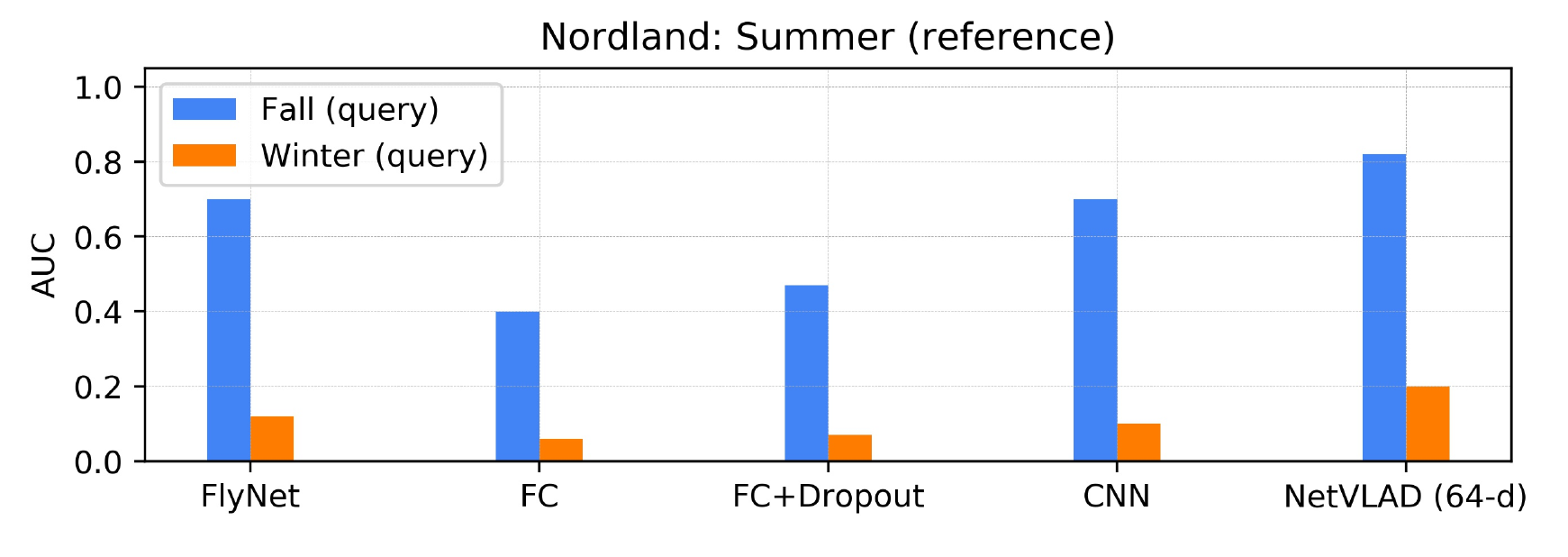}}
   \subfigure{
   \includegraphics[width=0.46\columnwidth]{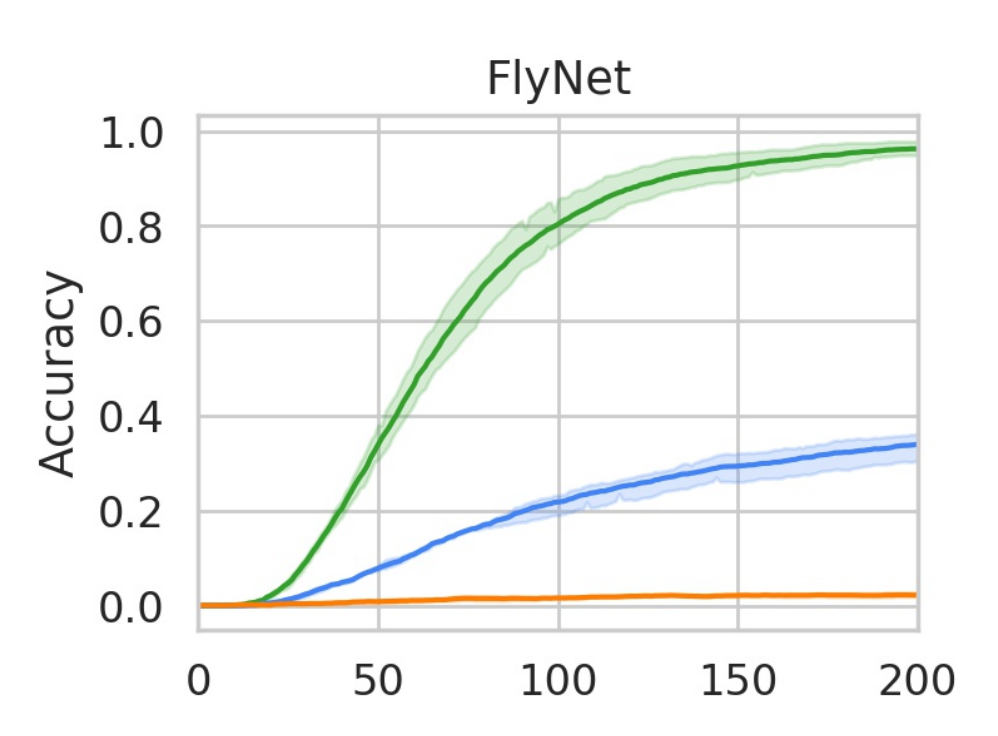}}
   \subfigure{
   \includegraphics[width=0.46\columnwidth]{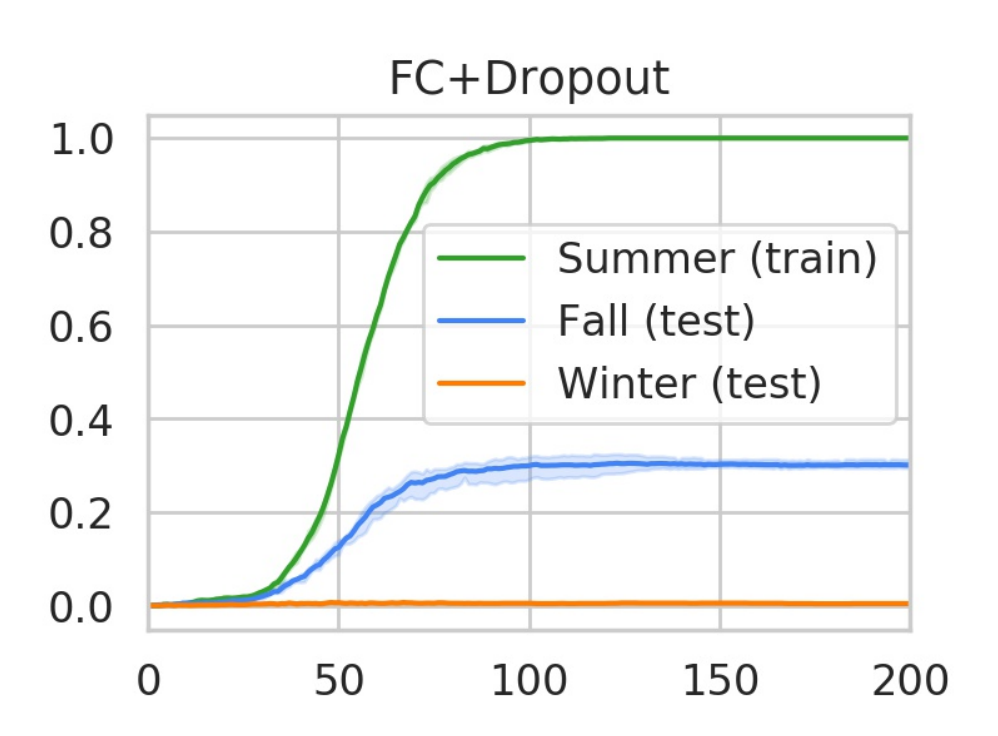}}
   \vspace{-2mm}
   \caption{\textbf{Comparison of FlyNet (alone) to other single-frame neural networks}. AUC results across different models on the Nordland dataset (left). Average accuracy over 10 training experiments vs. number of epochs for FlyNet (middle) and a fully connected (FC) network with dropout (right).}
   \label{fn_comparisons}
\end{figure*}

\begin{figure*}[!h]
   \vspace{-3mm}
   \subfigure{
   \includegraphics[width=\columnwidth]{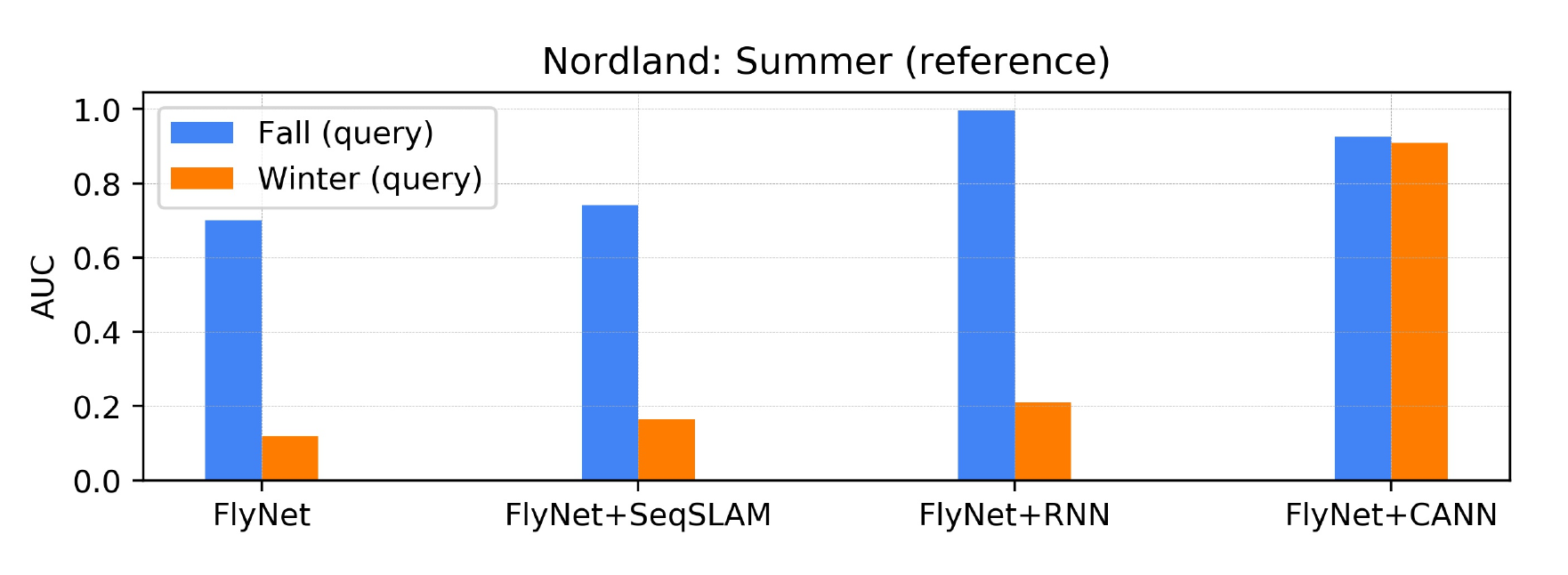}}
   \subfigure{
   \includegraphics[width=\columnwidth]{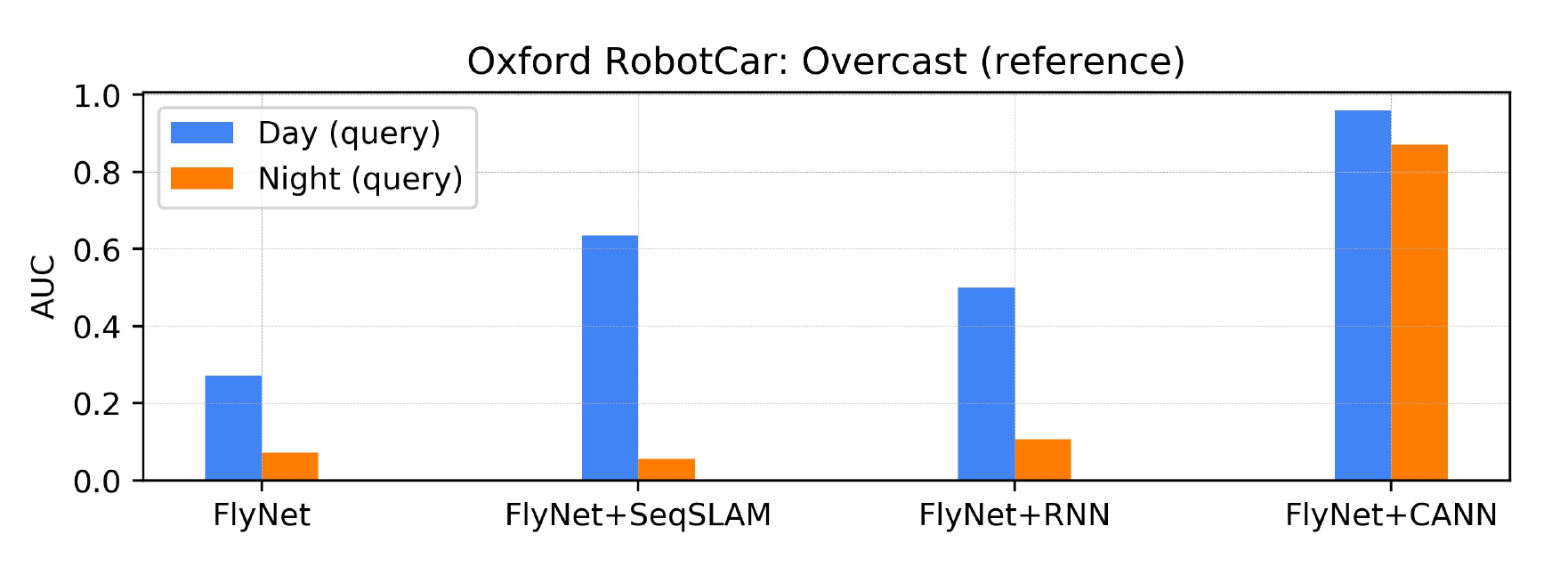}}
   \vspace{-2mm}
   \caption{\textbf{FlyNet baselines}. AUC results of single-frame and multi-frame FlyNet-based models on Nordland (left) and Oxford RobotCar (right) datasets.}
   \label{comp_baselines}
   \vspace{-2mm}
\end{figure*}

\subsection{Comparison of FlyNet to other Neural Networks}\label{compare}

We compare FlyNet (alone) with a range of related single-frame models including FC networks that use dropout \cite{drop2014}, a vanilla CNN model often used in visual navigation research \cite{complex,impala}, and the well-known NetVLAD method \cite{netvlad}. We train all these models end-to-end using a 1000-way \textit{linear} MLP classifier---except for both the off-the-shelf NetVLAD backbone and the FNA layer in FlyNet (as its sparse matrix $\mathbf{W}$ stays unchanged). Average accuracy results over ten experiments using different seed numbers are shown in Fig. \ref{fn_comparisons}.

For FlyNet, we use its FC output layer as the \textit{linear} classifier, as shown in Fig. \ref{flynet} (a). For the FC networks, we use a three-layer MLP with 64--64--1000 units respectively, as in the FlyNet architecture. We then obtain the FC+Dropout network by using dropout rates of 90\% and 50\% for the first and second layers of the FC model, respectively, in order to approximate the FlyNet sparsity and for fair comparison purposes. For the CNN model, we use 2 \textit{convolutional} layers but with gray-scale input images of $32\times64$ as in FlyNet. For NetVLAD, we use RGB images of 244$\times244$, as required by its off-the-shelf VGG-16 \cite{vgg} model, but we reduce their output representation dimensionality from 4096-\textit{d} to 64-\textit{d} to be comparable in size with the FlyNet representation. It is worth noticing that we do not reduce the CNN and NetVLAD model sizes down to the same size as FlyNet as they use pre-defined (rigid) architectures inherent to their approaches. We use the Adam optimizer \cite{adam} for training, and a learning rate set to 0.001 for all our experiments. \copyrightnoticetop \copyrightnotice

\subsection{FlyNet Baselines Experiments}

We trained and tested our four FlyNet baselines, described in Section \ref{sec:baselines}, in order to obtain our best performing model and compare it against existing state-of-the-art VPR methods. In Table \ref{table_params}, we show the number of layers, weights, and units for each model. For FlyNet and FlyNet+RNN, the FNA hidden layer used 64 units, and their FC layers used 1000 units, see Fig. \ref{flynet} (a, c). The number of recurrent units for FlyNet+RNN was 512. For FlyNet+CANN, the CANN layer used 1002 units. We also show the AUC performance of our FlyNet baselines on both the Nordland and Oxford RobotCar datasets in Fig. \ref{comp_baselines} to further analyze these results in Section \ref{sec:fn-baselines-eval}.

\begin{table}[!ht]
\caption{FlyNet Baselines Footprint}
\label{table_params}
\vspace{-2mm}
\begin{center}
\begin{tabular}{cccc}
\toprule
\textbf{Architecture} & \textbf{\# layers} & \textbf{\# params} & \textbf{\# neurons} \\
\midrule
FlyNet & 2 & 64k & 1064 \\
FlyNet+RNN & 4 & 1.3m & 2576 \\
FlyNet+CANN & 3 & 72k &  2066 \\
\bottomrule
\end{tabular}
\end{center}
\vspace{-4mm}
\end{table}

\subsection{Comparison to existing State-of-the-Art Methods}

We compare our best performing FlyNet-based model with the algorithmic technique SeqSLAM (without FlyNet attached), and two deep-learning-based methods: LoST-X and Multi-Process Fusion.

\subsubsection{\textbf{SeqSLAM}}
SeqSLAM \cite{seqslam} shows state-of-the-art VPR results under challenging visual appearance changes. We use the MATLAB implementation in \cite{n2013}, with a sequence length of 20 frames, a threshold of 1, and the remaining SeqSLAM parameters using its default values.

\subsubsection{\textbf{LoST-X}}
The multi-frame LoST-X pipeline \cite{garg2018lost} uses visual semantics to perform VPR over day/night cycles, with further development for opposing viewpoints in \cite{ijrr}. LoST-X uses the RefineNet model \cite{refinet}, a ResNet-101-based model, as its semantic feature encoder, which is pre-trained on the Cityscapes dataset \cite{cityscapes} for high-resolution segmentation.

\subsubsection{\textbf{Multi-Process Fusion (MPF)}}
MPF \cite{mpf} is also a multi-frame VPR technique. We use the VGG-16 network \cite{vgg} trained on Places365 \cite{places365} to encode the images and feed the MPF sequence-based dynamic algorithm.

\begin{figure*}[!ht]
    \vspace{-3mm}
    \centering
    \subfigure{
        \includegraphics[width=\columnwidth]{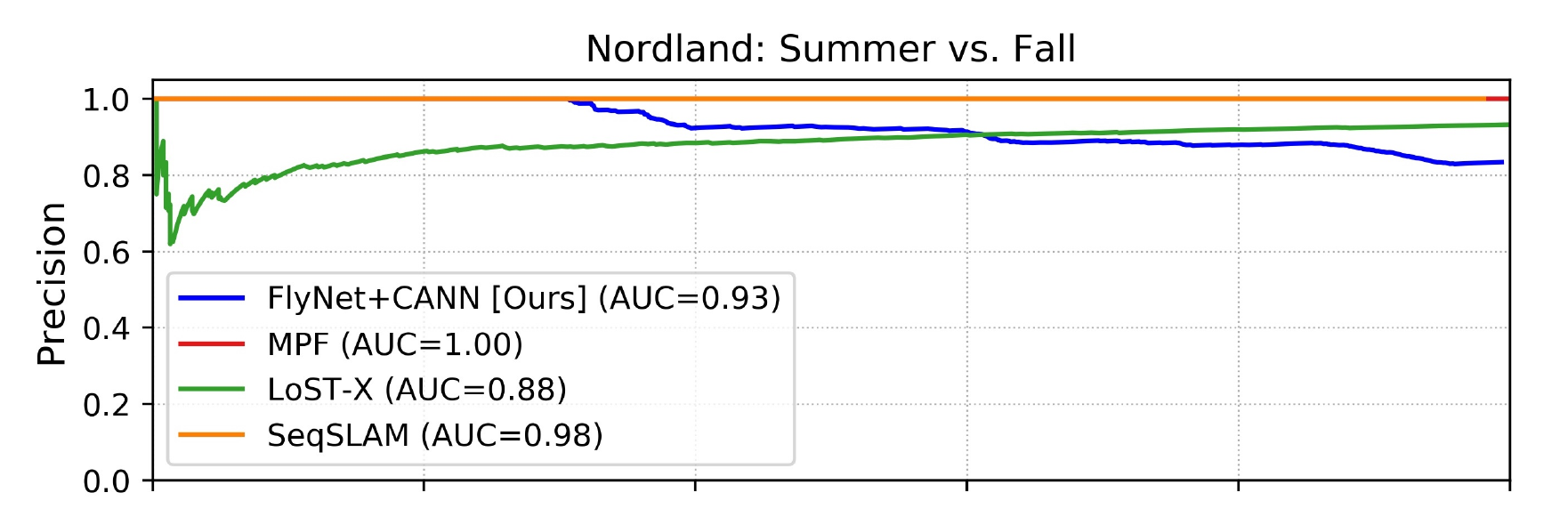}}
    \subfigure{
        \includegraphics[width=\columnwidth]{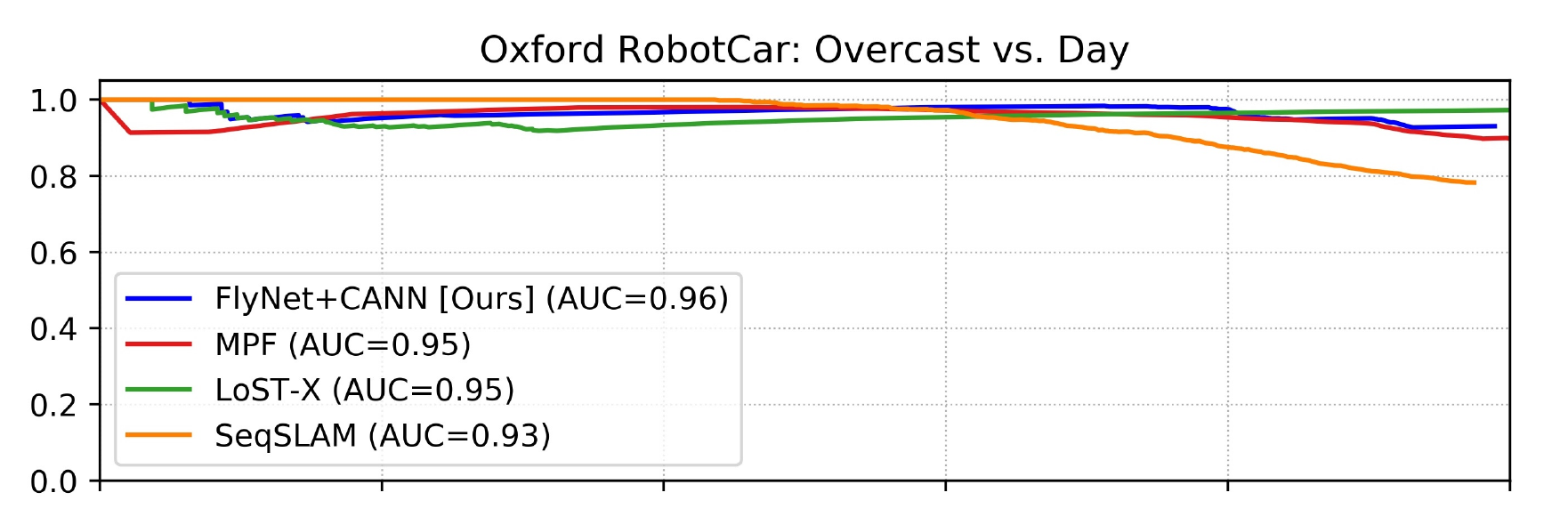}}
    \subfigure{
        \includegraphics[width=\columnwidth]{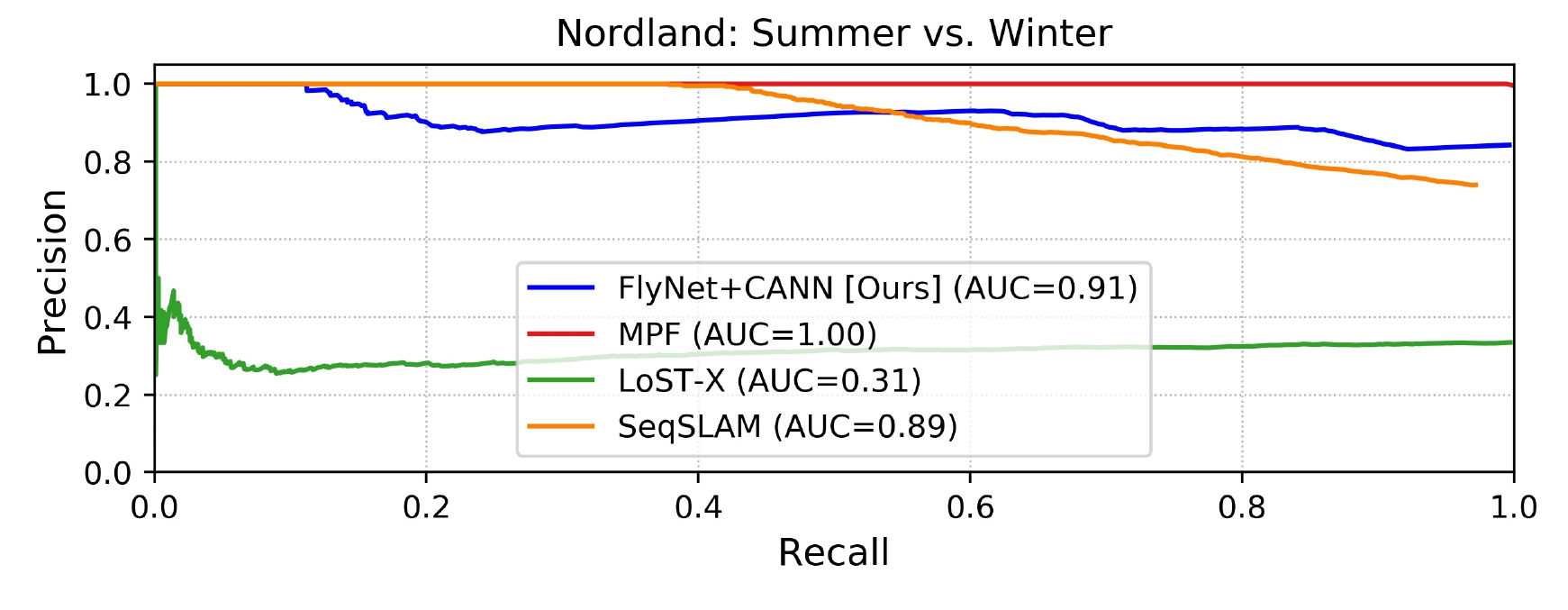}}
    \subfigure{
        \includegraphics[width=\columnwidth]{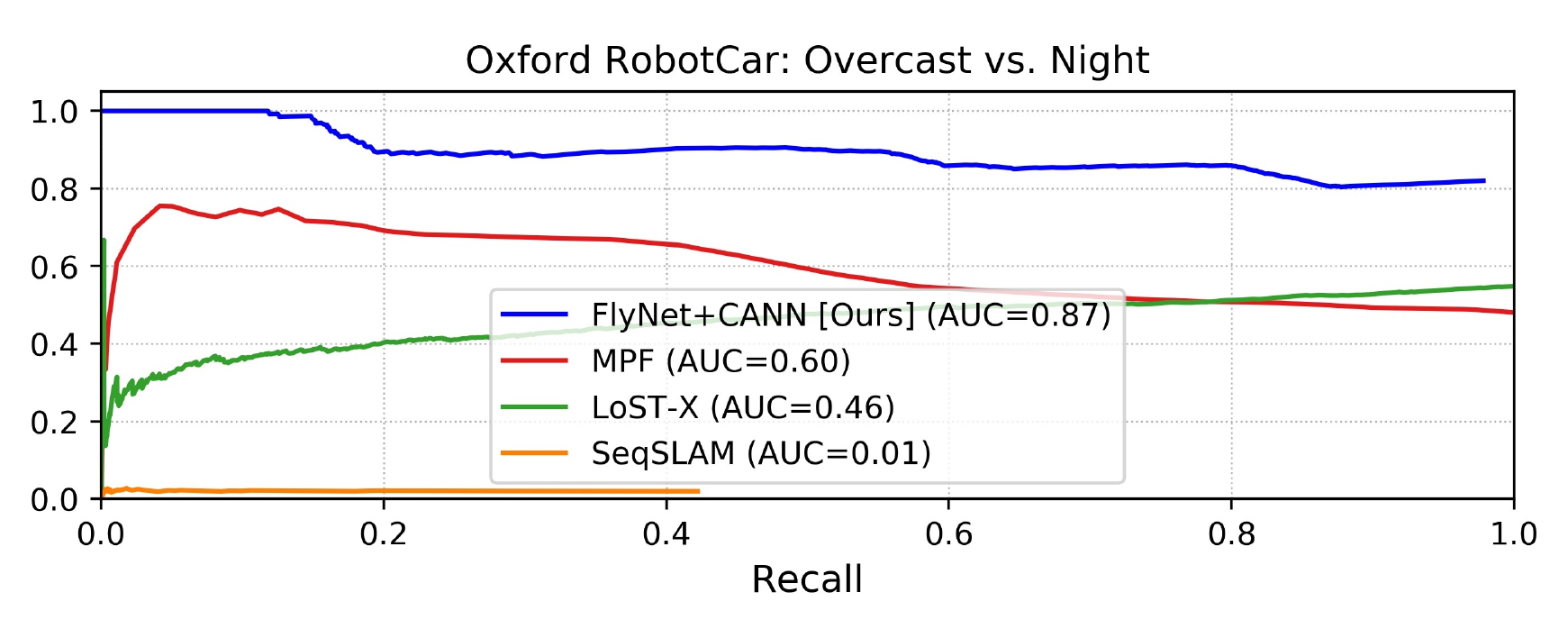}}
    \vspace{-8mm}
    \caption{\textbf{PR performance} of FlyNet+CANN vs. SeqSLAM, LoST-X and MPF on 1000-places of the Nordland (left) and Oxford RobotCar (right) dataset.}
    \label{pr}
    \vspace{-2mm}
\end{figure*}

\begin{figure*}[!h]
   \centering
   \subfigure{
       \includegraphics[width=\columnwidth]{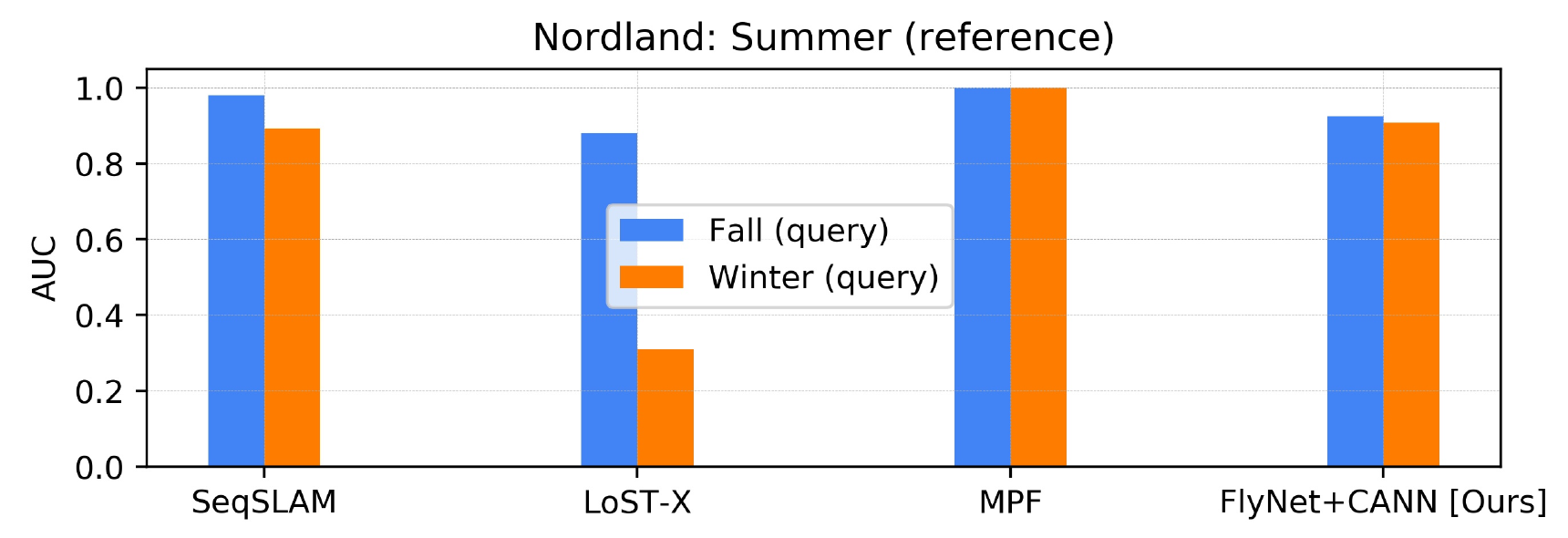}}
   \subfigure{
       \includegraphics[width=\columnwidth]{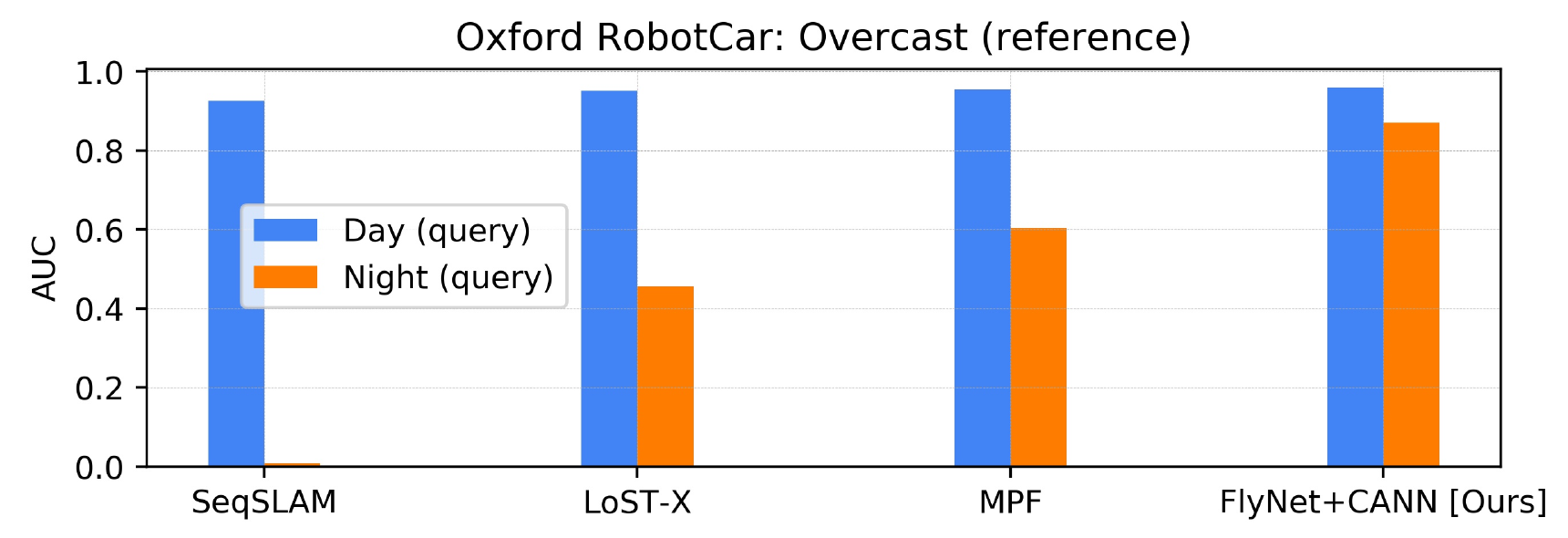}}
   \vspace{-8mm}
   \caption{\textbf{AUC results} of FlyNet+CANN compared to SeqSLAM, LoST-X, and MPF on the Nordland (left) and Oxford RobotCar (right) dataset.}
   \label{auc_no}
\end{figure*}

\section{Results}\label{sec:results}

In this section, we analyze the experiments shown in Section \ref{sec:experiments}, along with Figs. \ref{fn_comparisons} and \ref{comp_baselines}, and describe the results of PR curves and related AUC metrics for visual place recognition.

\subsection{FlyNet vs. other Single-frame Networks and VPR Models}\label{sec:fn_vs_single}

From Fig. \ref{fn_comparisons} (left), we can see that FlyNet is directly competitive with both FC networks, despite FlyNet having over 3 times fewer parameters (64k vs. 199k). Potentially using 32 times less memory, as the FNA layer require only 1-bit per binary weight, as per previous research \cite{bnn}, compared to the corresponding layer using 32-bit floating-point weights in the FC models. On the other hand, for the CNN and NetVLAD models, with 6 and 234 times more parameters than FlyNet respectively, the larger the model the better the results we obtained. Under small environmental changes (e.g. summer to fall) both networks achieved over 70\% AUC, comparable to FlyNet. However, under extreme visual changes (e.g. summer to winter) all these models show relatively similar results, below 12\% AUC, except for NetVLAD with 20\% AUC.

In Fig. \ref{fn_comparisons} (middle, right), we show in further detail the average training results of FlyNet against the FC model with dropout across 200 epochs. Additional experiments to support the choice of the FlyNet parameters (e.g. sampling ratio of 10\% and WTA circuit of 50\%) were also conducted. For the sampling ratio, we gradually increased it from 10\% to 90\% but no further accuracy improvement than 96\% was obtained. For the WTA parameter, we varied it between 5\% and 95\% but, as we moved away from 50\% WTA, the training accuracy decreased to 25\% and 40\%, respectively. \copyrightnoticetop \copyrightnotice

\subsection{FlyNet Baselines Evaluations}\label{sec:fn-baselines-eval}

Although there are significant performance differences at a single-frame matching level, Fig. \ref{comp_baselines} shows that when using sequence-based filtering techniques these differences reduce significantly. Meaning that using the more compact networks is viable in a range of applications where temporal filtering is practically feasible. It is possible then to leverage our compact FlyNet network and integrate it with a range of sequence-based methods such as SeqSLAM, RNN or CANN models and achieve competitive results. For FlyNet+SeqSLAM, the performance of FlyNet (alone) was significantly improved (see Fig. \ref{comp_baselines}). Similarly, the RNN layer on top of FlyNet improved even further these results. However, when integrating the output of FlyNet with a 1-\textit{d} CANN we were able to outperform these models, even under extreme environmental changes (e.g. day to night, summer to winter); we then choose this hybrid approach to compare against existing state-of-the-art methods.

\subsection{State-of-the-Art Analysis}

Figs. \ref{pr} and \ref{auc_no} show quantitative results for FlyNet+CANN and state-of-the-art VPR methods. Fig. \ref{pr} (left) shows the PR performance curves on the Nordland dataset, where MPF is performing better while being able to recall almost all places at 100\% precision on both fall and winter testing traverses. Achieving also the highest AUC results, see Fig. \ref{auc_no} (left). On the other hand, the semantic-based approach LoST-X can recall a few matches at 100\% precision on both testing traverses (fall and winter). In contrast, FlyNet+CANN achieves state-of-the-art results comparable with SeqSLAM and MPF in all these tested traverses, see Fig. \ref{auc_no} (left).

\begin{table}[!ht]
\caption{Processing time comparison on the Nordland dataset}
\label{table_time}
\begin{center}
\begin{tabular}{cccc}
\toprule
\textbf{VPR System} & \textbf{Feature Ext.} & \textbf{Place Match.} & \textbf{Avg. Time (fps)}\\
\midrule
\textbf{FlyNet+CANN} & \textbf{35 sec} & \textbf{25 sec} & \textbf{0.06 sec} (\textbf{16.66}) \\
MPF & 1.9 min & 4.6 min & 0.39 sec (2.56) \\
LoST-X & 110 min & 200 min & 18.6 sec (0.05) \\
SeqSLAM & 50 sec & 40 sec & 0.09 sec (11.11) \\
\bottomrule
\end{tabular}
\end{center}
\end{table}

\begin{figure*}[!t]
   \vspace{-3mm}
   \centering
   \subfigure{
   \includegraphics[width=\columnwidth,height=24mm]{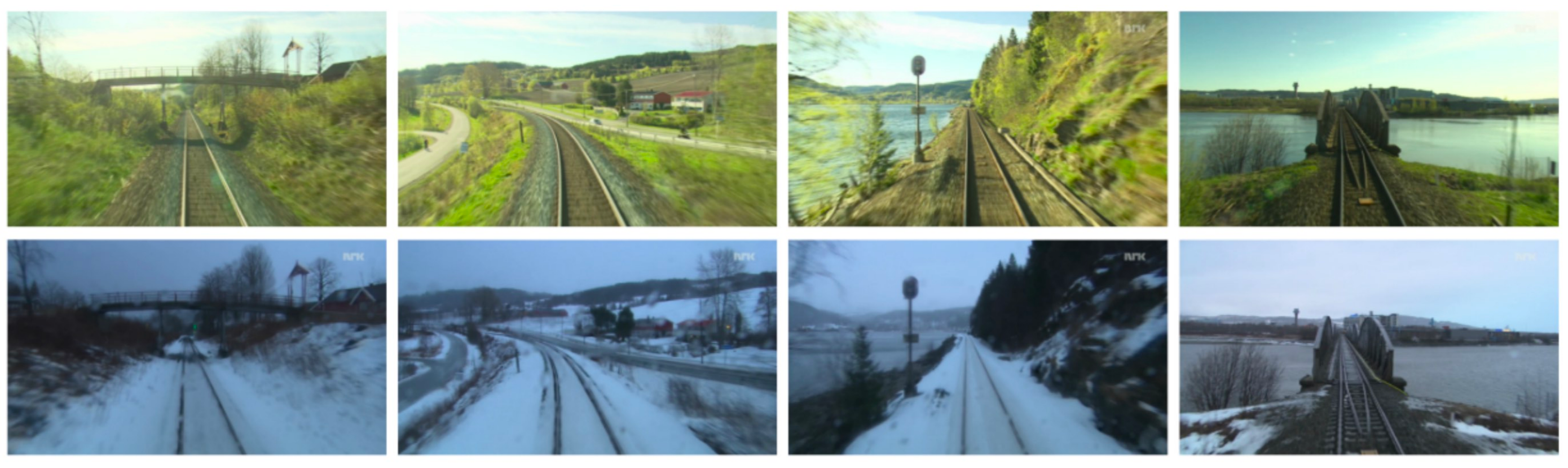}}
   \subfigure{
   \includegraphics[width=\columnwidth,height=24mm]{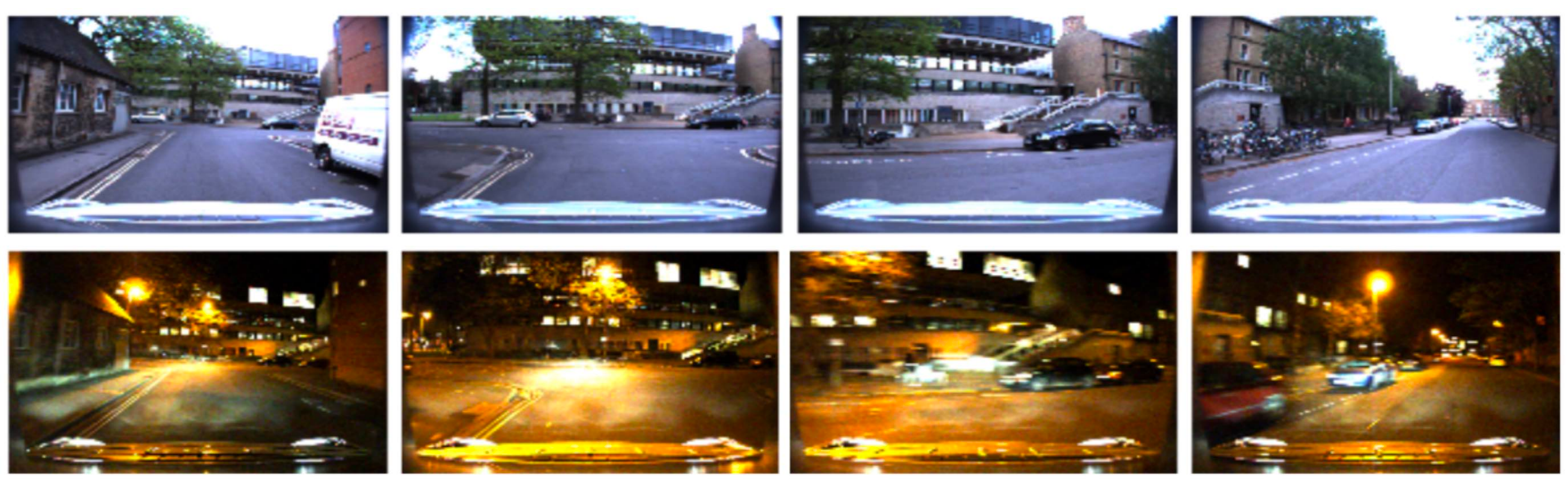}}
   \vspace{-6mm}
   \caption{\textbf{Generalization results}. Sample images (reference) of the Nordland summer (left-top) and Oxford RobotCar overcast traverses (right-top). Corresponding frames retrieved (query) using our FlyNet+CANN model from the Nordland winter (left-bottom) and Oxford RobotCar night traverses (right-bottom).}
   \label{seq-results}
   \vspace{-2mm}
\end{figure*}

Similarly, PR performance on the Oxford RobotCar dataset is shown in Fig. \ref{pr} (right). Also notable in this case is that FlyNet+CANN again achieves state-of-the-art results that are now comparable with SeqSLAM, LoST-X, and MPF approaches. Our hybrid model consistently maintains its PR performance even under extreme environmental changes (e.g. overcast to night), see Fig. \ref{pr} (right-bottom). In Fig. \ref{auc_no} (right), we also show how FlyNet+CANN outperforms the remaining methods in terms of AUC results, and Fig. \ref{seq-results} shows qualitative generalization results on both datasets. \copyrightnoticetop \copyrightnotice

\subsection{Computational Performance}

The processing time required to perform appearance-invariant VPR by our hybrid model is compared to those from state-of-the-art methods in terms of running time for (1) feature extraction, (2) visual place matching between query and reference traverses, and (3) average place recognition time for a single query image from a 1000-image reference database. This \textit{Avg. Time} \textit{(3)} is calculated as (\textit{Feature Ext. (1)} + \textit{Place Match. (2)})/1000. Processing time results on the Nordland dataset are reported in Table \ref{table_time}, where we show that our hybrid approach can be up to 6.5, 310, and 1.5 times faster than MPF, LoST-X, and SeqSLAM, respectively.

\begin{figure}[!t]
   \centering
   \includegraphics[width=\columnwidth]{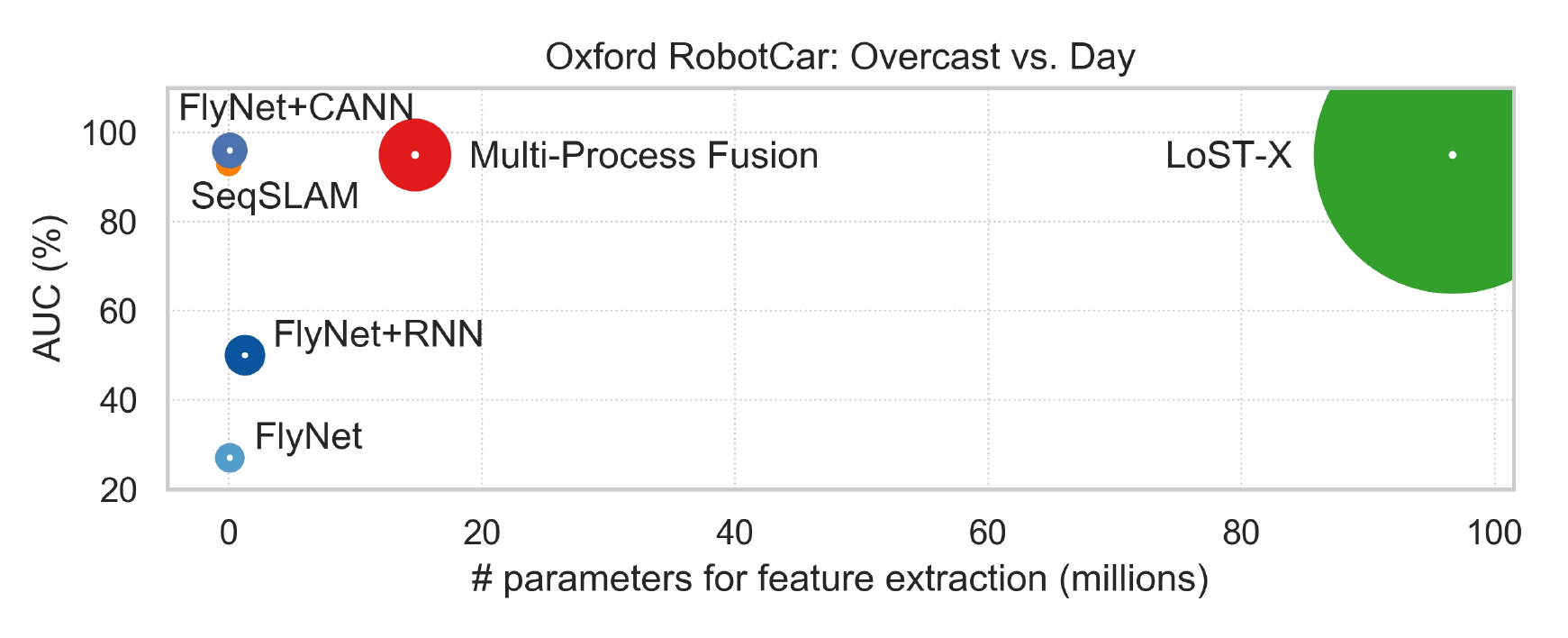}
   \vspace{-8mm}
   \caption{\textbf{Oxford RobotCar AUC performance vs. Model Size}. Similar to Fig. 1, it compares small appearance changes (overcast vs. day).}
   \label{model_size_2}
\end{figure}

Fig. \ref{model_size_2} shows a similar comparison presented in Fig. \ref{model_size_1} but with moderated appearance changes (overcast to day) on the Oxford RobotCar dataset. In this figure, again, the area of the circle is proportional to the number of layers per model, except for SeqSLAM which performs an algorithmic matching procedure. State-of-the-art methods like MPF, LoST-X, and SeqSLAM achieve better AUC results than in Fig. \ref{model_size_1} with 95\%, 95\% and 93\% respectively, where FlyNet+CANN also shows competitive results with 96\% AUC.

\subsection{Influence of Bio-inspiration}

In Figs. \ref{pr}--\ref{model_size_2} and Table \ref{table_time}, we show how our proposed FlyNet+CANN model achieves competitive visual localization results compared to existing deep learning and algorithmic-based VPR techniques, but with significantly fewer parameters, a smaller footprint and reduced processing time. Although we could have used conventional, pre-trained CNN models instead of FlyNet, our objective is to demonstrate to what extent we can draw inspiration from the brain's structural and functional connections between neural cells. Making possible to develop a sample-efficient, high-performing hybrid neural model, which structure is aligned with algorithmic insights found in the brain, as outlined in previous work \cite{whatnn, worm}, but for VPR tasks.

For FlyNet, it has the same number of layers and sparse structure found in the fly olfactory neural circuit. Although the fly brain expands the dimensionality of their input odor \cite{dasgupta2017}, e.g. from $m$ to $n=40m$ (see Fig. \ref{fna_pic}). We experimentally found that by reducing this dimension, e.g. from $m$ to $n=m/32$ instead, the FlyNet training accuracy remained around 96\%, as shown in Fig. \ref{fn_comparisons} (middle), while preserving the desired compact network structure.

For FlyNet+CANN, the integration of a 1-\textit{d} CANN model, for temporally filtering the output of FlyNet, enabled the use of a relatively low-performance but fast network to get better VPR results for our whole hybrid model, which is also able to generalize across challenging environmental changes (see Fig. \ref{seq-results}), while being up to three orders of magnitude faster than existing VPR methods, see Table \ref{table_time}.

\section{Conclusion}\label{sec:conclusion}

We proposed a new bio-inspired, hybrid model for visual place recognition based by part on the fruit fly brain and integrated with a compact neurally-inspired continuous attractor neural network. Our model was able to achieve competitive place recognition performance and generalize over challenging environmental variations (e.g. day to night, summer to winter), compared to state-of-the-art approaches that have much larger network size and computational footprints. It was also, to the best of our knowledge, the furthest in capability an insect-based place recognition system has been pushed with respect to demonstrating real-world appearance-invariant VPR without resorting to full deep learning architectures.

Future research bridging the divide between well-characterized insect neural circuits \cite{h16, b16} as well as recent deep neural network approaches and computational models of network dynamics related to spatial memory and navigation \cite{c18} are likely to yield further performance and capability improvements, and may also shed new light on the functional purposes of these biological neural networks. \copyrightnoticetop \copyrightnotice


%



\section*{Acknowledgment}

The authors thank Jake Bruce currently at Google DeepMind for insightful discussions about the potential ways to implement the FlyNet+RNN model, and also thank Sourav Garg, Stephen Hausler, and Ming Xu for helpful discussions.

\end{document}